\def\year{2018}\relax
\documentclass[letterpaper]{article} 
\usepackage{aaai18}  
\usepackage{times}  
\usepackage{helvet}  
\usepackage{courier}  
\usepackage{url}  
\usepackage{graphicx}  
\usepackage{amsmath}
\usepackage{mathtools}
\usepackage{algpseudocode}
\usepackage{algorithm}
\usepackage{multirow}
\usepackage{subcaption}

\usepackage{color}

\begin{document}
%
\title{Tracing Player Knowledge in a Parallel Programming Educational Game}
\author{
Pavan Kantharaju \and Katelyn Alderfer \and Jichen Zhu \\
{\bf \Large Bruce Char \and Brian Smith \and Santiago Onta{\~n}{\'o}n} \\
Drexel University \\
Philadelphia, PA 19104 \\
\{pk398, kmb562, jz465, charbw, bks59, so367\}@drexel.edu \\
}

\maketitle
\begin{abstract}
This paper focuses on {\em tracing player knowledge} in educational games. Specifically, given a set of concepts or skills required to master a game, the goal is to estimate the likelihood with which the current player has mastery of each of those concepts or skills. The main contribution of the paper is an approach that integrates machine learning and domain knowledge rules to find when the player applied a certain skill and either succeeded or failed. This is then given as input to a standard knowledge tracing module (such as those from Intelligent Tutoring Systems) to perform knowledge tracing. We evaluate our approach in the context of an educational game called {\em Parallel} to teach parallel and concurrent programming with data collected from real users, showing our approach can predict students skills with a low mean-squared error.
\end{abstract}

\section{Introduction}\label{sec:intro}
This paper focuses on the problem of {\em player/student knowledge modeling} in the context of educational games. Player knowledge modeling is the problem of estimating the am\-ount of knowledge or mastery that the current player possesses in a certain set of concepts or skill of interest. This problem has been studied in several fields, such as Game AI and Intelligent Tutoring Systems (ITS) (where it is known as ``knowledge tracing''~\cite{corbett1994knowledge,pavlik2009performance}). Specifically, in this paper we present a new knowledge modeling approach designed to monitor, in real-time, the degree of mastery that the current player has on the different skills required to play an educational game, based only on in-game player activity. A key contribution of this work is being able to perform knowledge modeling in complex educational games, where it is hard to determine when did a student apply a certain skill, and whether it was successful or unsuccessful.

Our approach is presented in the context of an educational game to teach parallel and concurrent programming called {\em Parallel}~\cite{ontanon2017designing}. {\em Parallel} is an adaptive game that presents different players with a different level progression, depending on their needs. In order for this adaptation to happen, the game needs to maintain an estimation of the knowledge of the current player (which concepts does the current player understand and which it does not). It then presents each player with levels that are generated via procedural content generation (PCG) based on this estimation. 



To address this problem, we build upon existing work on player modeling~\cite{yannakakis2013player}, and work from the Intelligent Tutoring Systems (ITS) community on knowledge tracing~\cite{corbett1994knowledge,pavlik2009performance}. One of the challenges we faced when attempting to integrate existing work on knowledge tracing into {\em Parallel is that knowledge tracing assumes that assessing whether students are successfully or unsuccessfully deploying certain skills is easy.} However, 
in real gameplay sessions students deploy skills in an interweaved manner while playing the game, making this assessment non trivial. For example, students might drag and drop different elements onto the game board just to explore the behavior of certain game elements, making it hard to assess when they deployed a skill correctly or incorrectly. 

Our main contribution integrates supervised machine lear\-ning inspired by existing work on player modeling and domain knowledge to assess successful skill application, and applies this assessment to knowledge tracing for educational games. Specifically, the problem outlined above is addressed by capturing live-telemetry data and using machine learning to make predictions concerning the {\em problem solving strategy} (e.g. ``trial and error'') students are currently deploying using time windows. This problem solving strategy is used as a proxy for whether they successfully or unsuccessfully understand the different skills involved in the current level. This is then fed to a knowledge tracing framework. To improve the accuracy of the model, we included a collection of domain-specific rules that complement the machine learning approach. We evaluated our approach using transcripts of several think-aloud user study sessions to manually generate ground truth with which to compare the predictions made by our model, showing a relatively low prediction error. Our results also show that the idea of predicting {\em problem solving strategy} from a series of time windows can be used to identify successful/unsuccessful applications of skills. 

\section{Background}\label{sec:related}

Two major areas of work are related to the work presented in this paper: player modeling and knowledge tracing. 
In a game environment, a player model is an abstracted description of a player capturing certain properties of interest such as preferences, strategies, strengths, or skills~\cite{van2003local}. Significant work exists in areas such as modeling player preferences in order to maximize engagement~\cite{Riedl2008,Thue2007}, providing better non-player-character AI \cite{Weber2009}, or game analytics~\cite{canossa2013}. 
For an overview on player modeling, readers are referred to existing surveys~\cite{Smith2011,Machado2011}.

The two approaches to modeling student knowledge most related to our work are \textit{Knowledge Tracing} (KT) and \textit{Performance Factor Analysis} (PFA). We refer  interested readers to~\citeauthor{harrison2012review}~\shortcite{harrison2012review} for an overview of these techniques and others used in Intelligent Tutoring Systems (ITSs)~\cite{sleeman1982intelligent}. 
One of the most common types of KT is Bayesian Knowledge Tracing (BKT)~\cite{corbett1994knowledge}, which uses Hidden Markov Models.  
The model parameters are trained using Expectation Maximization~\cite{dempster1977maximum}, Conjugate Gradient Search~\cite{corbett1994knowledge}, or brute-force search~\cite{brute_force_bkt}. 

One issue with Knowledge Tracing is that it assumes that there is a one-to-one mapping between questions and skills. However, in practice, questions usually require multiple skills to answer them correctly (levels in \textit{Parallel} require multiple interleaved skills). PFA~\cite{pavlik2009performance}, based on prior work on Learning Factor Analysis~\cite{cen2006learning}, addresses this issue by using a model independent of questions. Specifically, PFA models the performance of a student as follows:
\begin{align}
m(u,\mathcal{KC},c,n) &= \sum_{j \in \mathcal{KC}}(\beta_{j} + \gamma_{j}c_{u,j} + \rho_{j}n_{u,j} ) \label{eq1} \\
p_{u}(m) &= \frac{1}{1+ e^{-m}} \label{eq2}
\end{align} 
where the performance of a student $u$ is defined, given a set of skills $\mathcal{KC}$, as a function of the number of times ($c_{u,j}$) that the student has correctly applied a given skill $j$ or failed to apply it ($n_{u,j}$) in the past. The final performance of student ($p_u$) is then assessed using a logistic model (Equation ~\ref{eq2}). The model parameters ($\beta_j$ , $\gamma_{j}$ and $\rho_j$) can be trained using logistic regression. 
%
%
Thus, one necessary input to utilize these models is the assessment of successful or unsuccessful application of skills. This is not trivial in the context of educational games like {\em Parallel}, and is one of the contributions of this paper. We compare our work against a modification of PFA (described in our experimental evaluation section) for {\em Parallel}. Although it is possible to extend BKT for multiple-skill questions~\cite{gong2010comparing}, comparing against BKT is part of our future work.


\section{Parallel: A Game for Learning Parallel and Concurrent Programming Concepts}\label{sec:parallel}


\begin{figure}[tb]
    \centering
    \includegraphics[width=0.65\columnwidth]{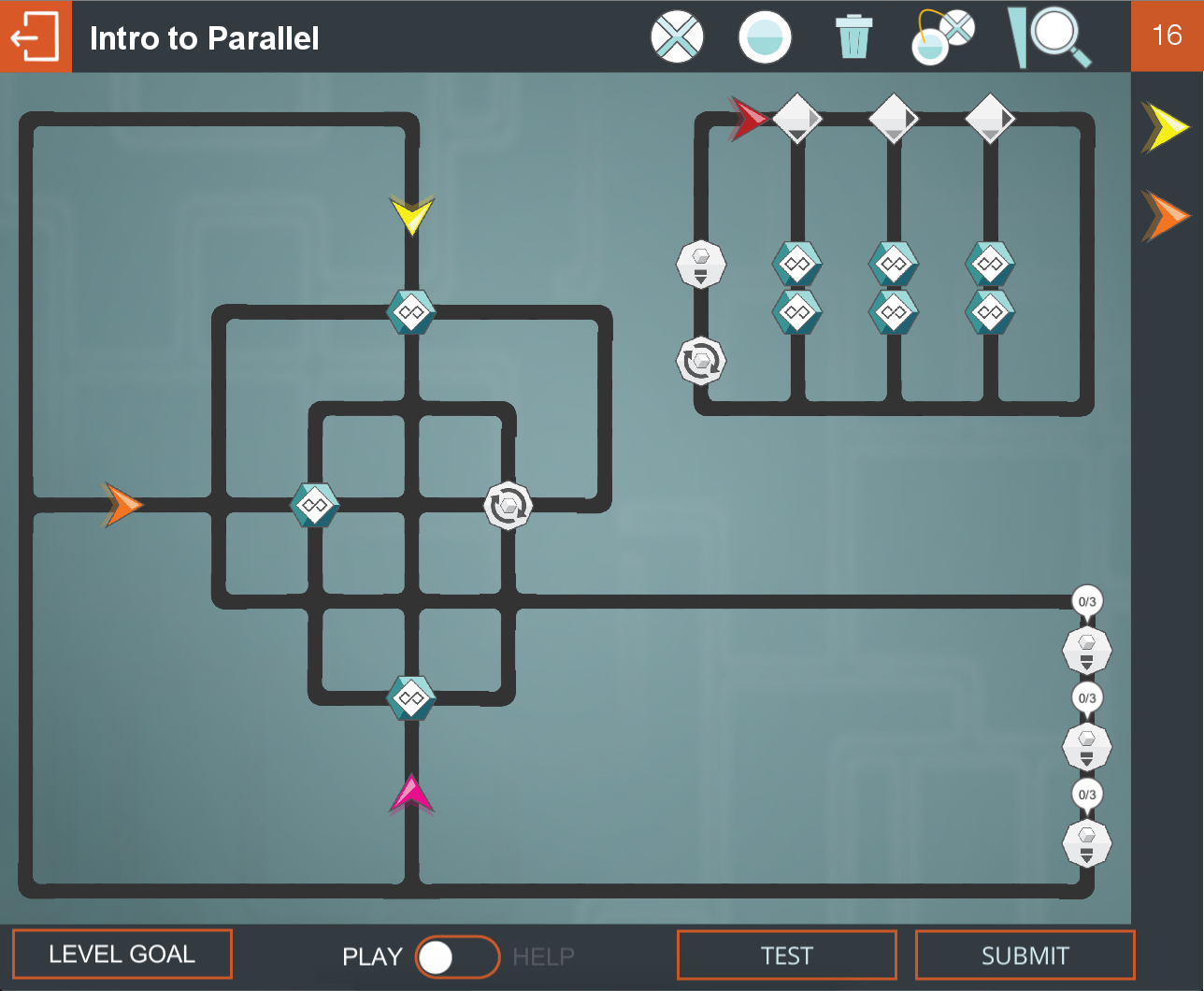}
    \includegraphics[width=0.65\columnwidth]{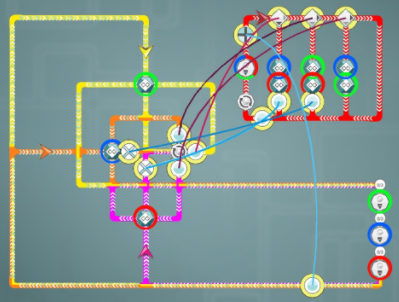}
    \caption{{\em Parallel}'s Visual Representation of the ``Cigarette Smokers Problem'' (top), and a possible solution (bottom), with colors used to highlight which arrows move through which tracks and their directions.}
    \label{fig:level16}
\end{figure}

\textit{Parallel}~\cite{ontanon2017designing} is an educational computer game designed to help students learn parallel and concurrent programming concepts as well as help us understand their learning processes. The game renders different parallel and concurrent programming concepts visually, and is designed as a puzzle game, where students must solve different problems in order to advance to the next level. Within each level (e.g., Figure \ref{fig:level16}), players see a collection of arrows that follow different tracks (black lines, representing programs). These arrows, which run at varying and unpredictable speeds (to represent non-determinism inherent in concurrent programming), represent threads. 
The main goal is to design synchronization mechanisms so that arrows accomplish the given challenge (e.g. deliver packages while preventing a ``race condition''). 

At any time while composing the solution to a level, a player can press the ``test'' button, which will make the game run a simulation of the current solution (with arrows moving stochastically). The solution might succeed or fail, but the fact that it succeeds does not mean that it is a correct solution, since it might be the case that if the arrows were to have moved at different relative speeds, the solution would fail. Students are free to ``test'' their solutions as many times as they want, before pressing the ``submit'' button, which causes a {\em model checker} to use systematic search to test every possible schedule of arrow movements, and see if the solution would work in all possible situations. If the solution works, the level is considered solved. 


The game presents problems of varying complexity in the form of different levels, from preventing simple race conditions, to solving classic situations such as the ``cigarette smokers problem'' (shown in Figure \ref{fig:level16}). 
Figure \ref{fig:level16} (bottom) shows a possible solution to this level with all the semaphores and signals in place and connected in the right way. Additionally, we colored the tracks with the color of the arrows that can go through them. Notice that solutions to levels like this are non-trivial, and would be very hard to find by trial and error. However, deploying concepts from parallel programming (such as the idea of ``identifying the critical section''), the solution is easier to find, and corresponds exactly to the typical solution to this problem in concurrent programming textbooks~\cite{downey2008little}. \textit{Parallel} has been deployed twice in a real undergraduate computer science course to help teach students about parallel programming.

\section{Player Knowledge Modeling in {\em Parallel}}\label{sec:student_knowledge_modeling}

\begin{figure}
\centering
\includegraphics[width=0.85\columnwidth]{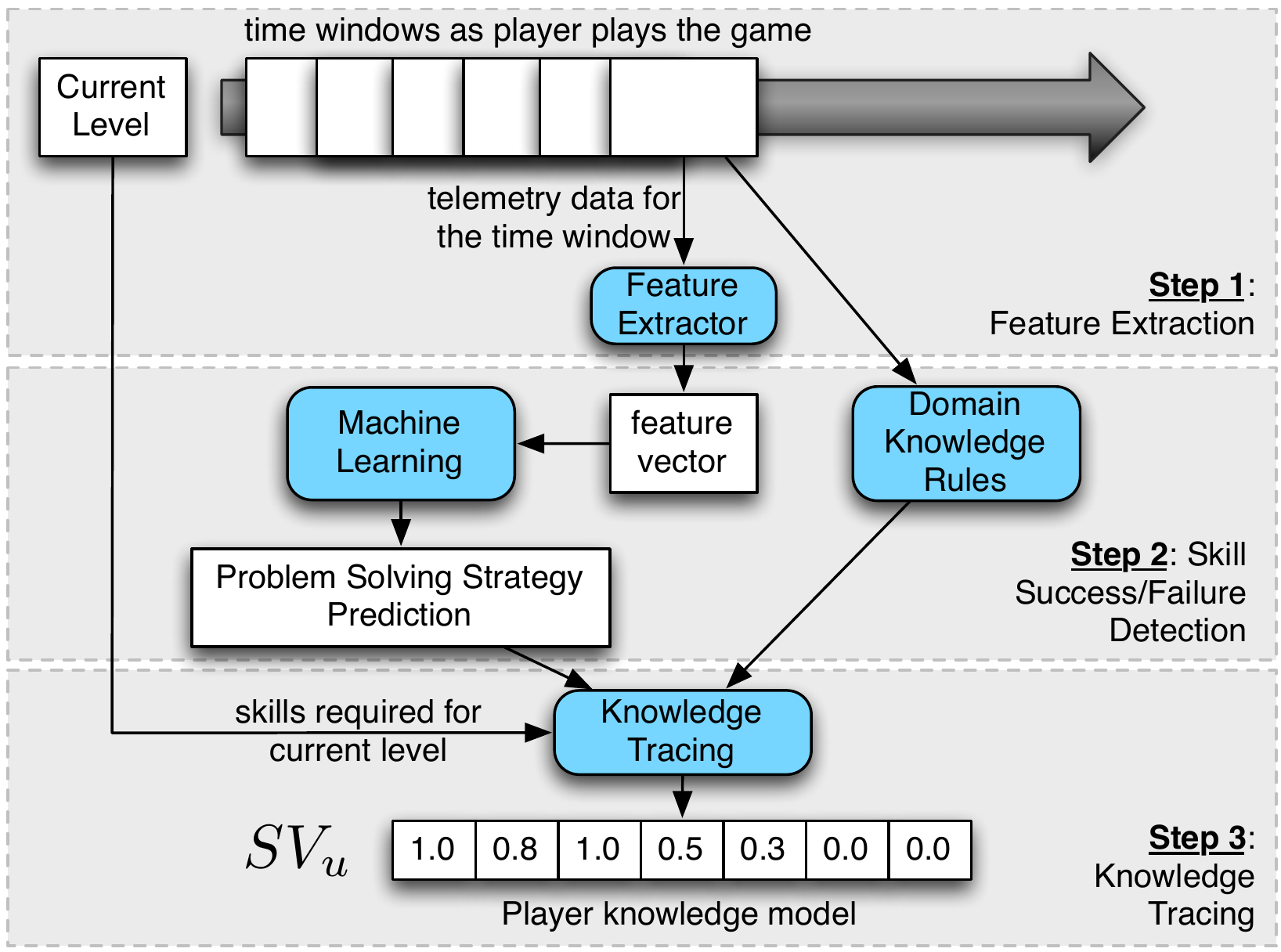}
\caption{Student Knowledge Modeling Process}
\label{fig:playerModelOverview}
\end{figure}

The goal of player modeling in {\em Parallel} is to generate the sequence of levels that a player will experience. Players should play game levels that help them practice the concepts they do not currently understand, and that are feasible given their current mastery of parallel programming. The output of our player modeling approach is a {\em player knowledge model} capturing the level of mastery of the current player in a set of concepts/skills required to play our game. In order to elicit the set of required concepts/skills, we employed the methodology defined by \citeauthor{horn2017adapting}~\shortcite{horn2017adapting}, which resulted in the 21 skills shown in the left column of Table \ref{tab:skills}, which we will denote by $\mathcal{KC}$.

Given $\mathcal{KC}$, the knowledge model of a given student $u$ is represented by a vector $SV_{u} = \langle p(s_{1}) , p(s_{2}) \dots p(s_{m}) \rangle$, where each element represents the student's understanding of each skill $s_{i} \in \mathcal{KC}$ as a likelihood that the student has mastered that skill (in the interval from 0 to 1).
A challenge in our domain is that, unlike in previous work on ITS, it is not obvious to identify when a player has attempted to deploy a skill, or when this deployment succeeded or failed. Suppose a level requires a student to apply two skills from Table~\ref{tab:skills} at the same time, \textit{Block critical sections} and \textit{Use diverters}, but the user is just focusing on making the arrows use the diverters in the right way. The fact that, for some time, they did not block the critical sections does not mean that they failed at applying the skill, but that they focused on something different. It is not trivial to infer what a player was focusing on, as they are free to place elements anywhere within the level and ``test'' the proposed solution at any time. 

In order to address this problem, our approach uses the idea of ``time windows'', and tries to identify successful/failed attempts of skill application in each time window. Specifically, it has 3 main steps (illustrated in Figure~\ref{fig:playerModelOverview}):
\begin{itemize}
\item {\em Step 1: Feature Extraction}: Real-time telemetry data concerning players actions are collected, and divided in time windows of fixed size. From each window, a collection of features is computed.
\item {\em Step 2: Skill Success/Failure Detection}: This step identifies whether a player was trying to successfully or unsuccessfully apply a certain skill. We employ two complementary approaches: a \textit{machine learning} component which predicts the {\em problem solving strategy} the player is currently deploying, from where successes or failures can be inferred as detailed below, and a collection of \textit{domain knowledge rules}, which can directly identify instances of skill application for some skills (shown in Table~\ref{tab:skills}).
\item {\em Step 3: Knowledge Tracing}: This step builds a knowledge model of a player using the output of the previous steps.
\end{itemize}
\noindent The remainder of this section describes each of these steps. 

\subsection{Step 1: Feature Extraction}\label{sec:step1}

Given a sequence of telemetry information of a student for a level, and a tunable time interval size $\tau$, we extract a feature vector based on the telemetry information from each time interval $(t,t+\tau)$ at intervals of $\tau/2$. The feature vector consisted of frequencies and rates (per minute) of different actions and events that occurred in game, as well as time differences between certain actions done by the player. We currently calculate 65 features.

\subsection{Step 2: Skill Success/Failure Detection}\label{sec:mltrain}


We use two complementary approaches to detect when players succeed or fail in deploying a skill: supervised machine learning and a collection of domain knowledge rules.

\noindent {\bf Machine Learning:} Given that it is hard to predict when students succeeded or failed in deploying skills in general, our approach instead predicts the {\em problem solving strategy} a student is deploying. After analyzing transcripts of think-aloud sessions from earlier user studies concerning how players play {\em Parallel}, we identified three basic problem solving strategies: {\em Trial and Error}, {\em Sequential Thinking}, and {\em Parallel Thinking}. We define \textit{Parallel Thinking} as the process of considering multiple arrows (threads) at the same time whereas \textit{Sequential Thinking} is defined as solving levels by considering one arrow at a time. \textit{Trial and Error} is the process of repeatedly trying different solutions to solve the current level by chance without any specific purpose. Given the set of skills in a given level $l$, $\mathit{KC}_l$, when our machine learning module predicts that the student is deploying trial and error for a given time window, we signal that there is a failed attempt at deploying all the concepts of $\mathit{KC}_l$ (notice that there might just be one of the concepts that has failed, but since we do not know which one, we signal them all). When we predict parallel thinking, we signal a successful application of the skills in $\mathit{KC}_l$ and for sequential thinking, we signal all skills with 0.5 probability of success. 
Notice the strong assumption in this process: when students deploy trial and error, we assume that they do not understand the logic behind a level. The high predictive accuracy reported in out experiments shows, however, that this assumption works in practice.


To train the machine learning model, we generate a training set from a set of annotated play-throughs. These annotations corresponding to problem solving strategies that were determined through analysis of one-on-one game-playing sessions with each student using a think-aloud protocol by a member of our research team (and revised by another). Standard supervised classification was used in our experiments, as described in the evaluation section. 

\begin{table*}[t]
\centering
\resizebox{\textwidth}{!}{%
\small
\begin{tabular}{|p{0.38\textwidth}|p{0.62\textwidth}|}
\hline 
\textbf{Skills}                                        & \textbf{Rules}                                                                                                                                                                                                                \\ \hline \hline
Hover over components to see what they do              & Player hovers over component                                                                                                                                                                                                  \\ \hline
Use help bar                                           & Player click on help bar and reads one or more of the guides                                                                                                                                                                  \\ \hline
Drag objects                                        & Player clicks and drags either semaphore or signal                                                                                                                                                                            \\ \hline
Place objects on the track                          & Player either places a semaphore or signal on track                                                                                                                                                                           \\ \hline
Hover over side arrows to see different colored tracks & Player hovers over arrows on side or clicks the side arrows                                                                                                                                                                   \\ \hline
Remove unnecessary elements                          & Player drags semaphore or signal to trash                                                                                                                                                                                     \\ \hline

Deliver packages                                       & All required packages are delivered                                                                                                                                                                                           \\ \hline
Be able to link signals to direction switches          & Player links a signal to a direction switch                                                                                                                                                                                   \\ \hline
Be able to link semaphores to signals                  & Player links a semaphore to a signal
\\ \hline
Understand the use of semaphores                       &                                                                                                                                                                                                                               \\ \hline
Understand that arrows move at unpredictable rates     & Player doesn't place multiple semaphores along one track without connecting to anything or doesn't move semaphore significantly.
\\ \hline
Understand that events happen in different orders     & 
\\ \hline
Use diverters                                          &                                                                                                                                                                                                                               \\ \hline
Prevent starvation                                     &                                                                                                                                                                                                                               \\ \hline
Block critical sections                                & Player places semaphore and signals in the proper positions to block critical section                                                                                                                                           \\ \hline
Synchronized multiple arrows                           & Player places semaphores and signals alternately on the tracks of the different arrows (a signal in arrow A's path is linked to the semaphore in arrow B's path, and vice-versa).
\\ \hline
Alternating access with semaphores and signals                             &                                                                                                                                                                                                 \\ \hline
Testing before submitting                              & Player tests before submitting                                                                                                                                                                                                \\ \hline
Understand specific delivery points                    & Packages are delivered correctly without losing any                                                                                                                                                                           \\ \hline
Understand exchange points                             & Packages are transferred at delivery points                                                                                                                                                                                   \\ \hline
Deliver packages with multiple synchronized arrows     &                                                                                                                                                                                                                               \\ \hline
\end{tabular}%
}
\caption{Rules used to detect successful application of each of the skills required to master {\em Parallel}.}
\label{tab:skills}
\end{table*}

\noindent {\bf Domain Knowledge Rules:} Additionally, a collection of manually defined domain knowledge rules are applied to each time window in order to detect additional evidence of successful or failed skill application. Table~\ref{tab:skills} shows the rules used in our experiments. Each rule is used to detect a certain skill using both telemetry information and information from the game's model checker, and was hand-authored by observing video recordings of students playing the game. 
Every time a rule is successfully fired, we signal a successful application of the corresponding skill.

\subsection{Step 3: Knowledge Tracing}\label{sec:knowledge_modeling}

Next, we describe the process of updating the student knowledge model using the output of machine learning, and the domain knowledge rules. Recall that time windows of telemetry data are generated while a player is playing a level. 
Let $F$ be the set of time windows generated for the current play-through of a player $u$ (which might include one or more levels), and let us denote by $\mathit{ML}(f)$ to the output of the machine learning classifier for time window $f$, encoding {\em trial-and-error} as 0, {\em single-threaded thinking} as 0.5, and {\em parallel thinking} as 1. Let us now call $R_{s_i}(f)$ to be the number of times domain knowledge rule $R_{s_i}$ was applied to time window $f$. Also, let $I_{s_i}(f)$ be the indicator function that is 1 if skill $s_i$ was involved in the level from where $f$ was extracted, and 0 otherwise. With these definitions, let us now define $R_{s_i}(F) = \sum_{f \in F} R_{s_i}(f)$, $I_{s_i}(F) = \sum_{f \in F} I_{s_i}(f)$, and $ML_{s_i}(F) = \sum_{f \in F | I_{s_i}(f) = 1} \mathit{ML}(f)$. Now, for a given student $u$, we compute the understanding of each skill $s_i \in \mathcal{KC}$ in the knowledge model $SV_{u}$ as follows:
\begin{align}
p(s_{i}) = \frac{ML_{s_i}(F) + R_{s_{i}}(F)} {I_{s_i}(F) + R_{s_{i}}(F)} \label{eq3}
\end{align}
\noindent Notice this is the average of the predictions made for a given skill by machine learning and the domain knowledge rules. 

\section{Experimental Evaluation}\label{sec:experiments}
\begin{table*}[tb]
\centering
\resizebox{\textwidth}{!}{%
\begin{tabular}{|c||c|c|c|c|c|c|c||c|}
\hline
Time Interval (sec) ($\tau$) & AdaBoost & Bagging & Bayes Net & J48    & Multilayer Perceptron & Naive Bayes & Random Forest  &  \textbf{Average} \\ \hline
10                  & {\bf 0.477}    & 0.4246  & 0.459     & 0.4393 & 0.3754                & 0.3361      & 0.423      & 0.4192  \\ \hline
20                  & {\bf 0.5299}   & 0.4602  & 0.4343    & 0.4299 & 0.4502                & 0.3361      & 0.494    &  0.4478   \\ \hline
30                  & 0.5356   & 0.5057  & {\bf 0.5563}    & 0.4044 & 0.4575                & 0.3287      & 0.5011   &   0.4699  \\ \hline
\end{tabular}%
}
\caption{Experiment 1: Classification Accuracy for Predicting Problem Solving Strategies (Dataset A)}
\label{tab:loo_accuracy}
\end{table*}

This section details our experimental evaluation including datasets, setup, and results. Concerning the machine learning module, we evaluated seven different machine learning techniques provided by the WEKA~\cite{weka2018} framework: \textit{J48}, \textit{Random Forest}, \textit{Bagging}, \textit{AdaBoost}, \textit{Naive Bayes}, \textit{Bayes Net}, and \textit{Multilayer Perceptron}. All machine learning algorithms were tuned to their default parameters. Additionally, we compared the performance of our approach against PFA~\cite{pavlik2009performance}.

\subsection{Datasets and Ground Truth} 

All datasets consist of telemetry logs from play sessions of \textit{Parallel} by real undergraduate students. Each log contains mouse movements and events triggered by the player. 

\noindent {\bf Datasets:} In our experiments, we use two different datasets. The first dataset, which we refer to as \textit{Dataset A}, contains data from 31 levels and was gathered from one-on-one sessions with six students where we asked them to think-aloud their thinking process as they played six levels in our game (not all students completed all levels). 
The second dataset, which we call \textit{Dataset B} contains data from 395 levels collected from an undergraduate parallel programming course of 17 students. For four weeks, each student was required to play some set of levels (different from Dataset A) each week at home. The set of levels played in datasets A and B were different.

\noindent {\bf Ground Truth:} 
The ground truth for Dataset A (both for problem-solving strategies, and for the level of mastery of each of the skills in the game), was manually generated by researchers on the team from transcripts of the think-aloud sessions and recorded videos of their gameplay. For each student, we calculated ground truth values for all skills for all the levels they played. For Dataset B, we requested each student in the undergraduate class fill out a survey assessing how well they understood each parallel programming concept each week, which is used as the ground truth. Since Dataset B has no think-aloud transcripts, no ground truth on problem-solving strategy exists for Dataset B, and thus, it cannot be used for training, but only to test our approach.

\subsection{Experimental Setup}
\begin{table*}[tb]
\resizebox{\textwidth}{!}{%
\begin{tabular}{|c||c|c|c|c|c|c|c||c|}
\hline
                    & \multicolumn{7}{c||}{\textbf{Machine Learning}}                                                 & \textbf{Rules} \\ \hline
Time Interval (sec) ($\tau$) & AdaBoost & Bagging & Bayes Net & J48     & Multilayer Perceptron & Naive Bayes & Random Forest & -           \\ \hline
10                  & 0.1383   & 0.1124  & 0.1383    & 0.1016  & 0.1383                & 0.2146      & 0.1010        & 0.1244          \\ \hline
20                  & 0.1383   & 0.1346  & 0.1383    & 0.3800  & 0.1377                & 0.2043      & {\bf 0.1064}      & 0.1244          \\ \hline
30                  & 0.1383   & 0.1321  & 0.1383    & 0.0917 & 0.5883                & 0.1195      & 0.1136    & 0.1244        \\ \hline
\end{tabular}
}
\resizebox{0.905\textwidth}{!}{%
\begin{tabular}{|c||c|c|c|c|c|c|c|}
\hline
                    & \multicolumn{7}{c|}{\textbf{Machine Learning + Rules}}                                                 \\ \hline
Time Interval (sec) ($\tau$) & AdaBoost & Bagging & Bayes Net & J48    & Multilayer Perceptron & Naive Bayes & Random Forest  \\ \hline
10 & 0.1383 & 0.1137 & 0.1383 & {\bf 0.0928} & 0.1383 & 0.1833  & 0.1025  \\ \hline
20 & 0.1383 & 0.1349 & 0.1383 & 0.2972  & 0.1378 & 0.1624  & 0.1086 \\ \hline
30 & 0.1383 & 0.1328 & 0.1383 & {\bf 0.0811} & 0.4244 & 0.0981 & 0.1162 \\ \hline
\end{tabular}%
}
\caption{Experiment 2: MSE for Estimating Student Skill (Dataset B)}
\label{tab:mse_8}
\end{table*}

We had three objectives for our experimental evaluation. First, we wanted to analyze the accuracy of machine learning techniques in predicting problem solving strategies (Experiment 1). Second, we wanted to evaluate the performance of combining machine learning and domain knowledge for player knowledge modeling (Experiments 2 and 3). Finally, even if some of the information required to execute Performance Factor Analysis (PFA) is not directly available in our setting, we wanted to compare against PFA assuming that this information were available (Experiment 4).

\noindent {\bf Experiment 1:} We computed prediction accuracy for each machine learning classifier in predicting problem solving strategy over three time intervals $\tau$ (shown in Table~\ref{tab:loo_accuracy}) using leave-one-student-out cross validation on the 31 traces from Dataset A. We tested on data from one student and trained on data from the remaining. 

\noindent {\bf Experiments 2 and 3:}  For the next two experiments, we compared the use of both machine learning and rules for knowledge tracing against several baselines: random prediction, 
``always predict 1'' (since skill values of 1.0 are the most common in the ground truth, this performs better than random), 
using only machine learning (ML), and using only rules (R). For experiment 2, we trained on Dataset A and tested on Dataset B, and for experiment 3, we ran a leave-one-out cross validation on the set of students using Dataset A. This leave-one-out policy is the same as the one employed in Experiment 1. Since predictions of skill value are numbers between 0 and 1, we used the Mean-Squared Error (MSE) as a measure of error (lower is better).

\noindent {\bf Experiments 4:}  We slightly modified PFA to effectively compare against our approach. Normally, we would estimate the parameters $\beta_{j}$, $\gamma_{j}$ and $\rho_{j}$ for some skill $j$ using logistic regression classification. In our case, we assumed $\beta_{j}$ remains constant (a common assumption in later PFA work), and learn the parameters $\gamma_{j}$ and $\rho_{j}$. We also compute performance for each skill. Thus, Equation~\ref{eq1} becomes:
\[m_{j}(u,\mathcal{KC},c,n) = \sum_{j \in \mathcal{KC}}(\gamma_{j}c_{u,j} + \rho_{j}n_{u,j})\]
We note that training data must be binary for training PFA (via logistic regression). We used the ground truth from Dataset A because much of the ground truth values were binary. Any ground truth that was not binary was discarded during training. We also note that $c_{u,j}$ and $n_{u,j}$ were computed using the binary values from the ground truth of Dataset A. Thus, notice that performance reported for PFA assumes ideal conditions were this information is available (it is not available in realistic conditions in our domain), and thus performance reported for PFA should be taken as an upper bound on the performance we can expect to achieve.

\subsection{Results}

\begin{table*}[tb]
\resizebox{\textwidth}{!}{%
\begin{tabular}{|c||c|c|c|c|c|c|c||c|}
\hline
                    & \multicolumn{7}{c||}{\textbf{Machine Learning}}                                        & \textbf{Rules} \\ \hline
Time Interval (sec) ($\tau$) & AdaBoost & Bagging & Bayes Net & J48    & Multilayer Perceptron & Naive Bayes & Random Forest & -              \\ \hline
10                  & 0.1098   & 0.2121  & 0.2109   & 0.1597 & 0.1975                & 0.2856      & 0.2199     	& 0.1138         \\ \hline
20                  & 0.1098   & 0.1766  & 0.2271   & 0.1439 & 0.1803                & 0.3037      & 0.1882      & 0.1138         \\ \hline
30                  & 0.1098   & 0.1900    & 0.2156   & 0.1559 & 0.1814                & 0.1920       & 0.1567      	& 0.1138         \\ \hline
\end{tabular}%
}
\resizebox{0.93\textwidth}{!}{%
\begin{tabular}{|c||c|c|c|c|c|c|c|}
\hline
                    & \multicolumn{7}{c|}{\textbf{Machine Learning + Rules}}                                                 \\ \hline
Time Interval (sec) ($\tau$) & AdaBoost & Bagging & Bayes Net & J48    & Multilayer Perceptron & Naive Bayes & Random Forest \\ \hline
10 	& {\bf 0.0938} 	& 0.1572 	& 0.1556 	& 0.1101 	& 0.1434 	& 0.1954 & 0.1598 \\ \hline
20 	& {\bf 0.0938} 	& 0.1245 	& 0.1619 	& 0.1033 	& 0.1264 	& 0.1967 & 0.1340   \\ \hline
30 	& {\bf 0.0938} 	& 0.1396 	& 0.1550  	& 0.1078 	& 0.1193 	& 0.1284 & 0.1079 \\ \hline
\end{tabular}%
}
\caption{Experiment 3: MSE for Estimating Student Skill (Dataset A)}
\label{tab:mse_35}
\end{table*}

{\bf Experiment 1:} Table~\ref{tab:loo_accuracy} provides accuracy measures for different classifiers over different time intervals. 
Overall, we see that larger time intervals achieve better results (seen in the {\em average} column on the right-hand side). The best performance was achieved using a Bayes Net with $\tau = 30$ seconds (55.63\%). Notice this is a 3-way classification problem, so, baseline classification accuracy of a random predictor would be 33\%. Thus, there is definitely a signal in our dataset that can be used for player modeling. 

\noindent {\bf Experiment 2:} Table~\ref{tab:mse_8} provides the MSE over all skills for Dataset B for machine learning only (ML), rules only (R), and with machine learning with our rules (ML+R). We note that the random baseline has an MSE of 0.25, and the ``always predict 1" baseline has an MSE of 0.1383. Most of the classifiers (except Multilayer Perceptron using a 30 second time interval) were able to beat the baselines. We see that AdaBoost and Bayes Net's MSE was constant for all time intervals for ML, R, and ML+R. This was due to both classifiers predicting each skill in a student's skill vector as 1.0.
We also note that MSE for R remained constant over all time intervals, implying that the time interval doesn't influence our rules. This makes sense because skills are detected from the telemetry information. The lowest MSE for ML and ML+R was J48 with a 30 second time interval (0.0917 for ML and 0.0811 for ML+R), beating R with an MSE of 0.1244. Looking closer at the MSE for each skill, we notice that the largest difference between ML and ML+R over each skill was 0.04 for ``Place objects on the track." With fine tuning of our rules, we can expect better performance. 

\noindent {\bf Experiment 3:} Table~\ref{tab:mse_35} provides the MSE over all skills for Dataset A for ML, R,  and ML+R. The lowest MSE for ML was AdaBoost with 0.1098, for ML+R was AdaBoost with 0.0938, and R with 0.1138. We see that every classifier for ML+R and most of the classifiers for ML outperformed the random baseline. However, in this dataset, none of the classifiers for ML outperformed the ``always predict 1" baseline, which was 0.0895 for this dataset. For ML+R, AdaBoost was the only classifier that was close to this baseline with an MSE of 0.0938. This suggests that, for this dataset, it is better to just always predict that a student knows all the skills with likelihood 1.0. The ground truth annotations were provided by hand by researchers in our team, and only those annotations for which two separate researchers were confident were kept. Because of this, most annotations are for 1.0 values (in our game it's easier to assess that a student knows something than the opposite). This makes this dataset skewed, and results on it concerning student skill prediction not very meaningful, compared to those from Dataset B where the ground truth was annotated directly by the students. However, we included them for completeness. 

\noindent {\bf Experiment 4:} We use the values from ML+R in Table~\ref{tab:mse_35} to compare against PFA. PFA could not issue predictions for all the skills, since for some of them there was no ground truth that was either 0 or 1, and thus we could not train the model using logistic regression. Assuming baseline 0.5 predictions for those skills, PFA achieves an error of 0.0655, lower than the best results we obtained (0.0938). Over the skills in which it could make predictions, PFA achieved an MSE of just 0.0450. Recall that in order to make PFA applicable, we feed part of the ground truth as part of its input, which would not be available in realistic conditions. Thus, this gives us a lower bound on the MSE that we can expect to achieve if our approach perfectly identified all instances of successful and unsuccessful skill application. An immediate line of future work is to replace Equation \ref{eq3} in our approach by PFA, which we expect will significantly improve results.
 
\noindent {\bf Discussion: } Our experimental evaluation sho\-ws that our pro\-posed approach outperforms the baselines in extracting information that is useful for knowledge tracing via the combination of machine learning to predict problem solving strategy and domain knowledge rules (which performs better than either of the two approaches in isolation). We notice that different machine learning techniques provide better results under different time intervals. This might be due to the difference in feature vectors at the given time intervals.

\section{Conclusions}\label{sec:conclusion}

This paper presents an approach to player knowledge tracing for an educational game. Our approach is based on integrating machine learning and domain knowledge rules that indicate when players successfully or unsuccessfully apply skills, and using those predictions to perform knowledge tracing. 
Our empirical analysis with data from real users shows that we can predict a student's understanding of skills with relatively low Mean-Squared Error: much lower than the baselines, and very close that achieved by PFA in an idealized situation were ground truth of the successful or unsuccessful application of skills was available.
As part of our future work, we would like to expand our domain knowledge rules and 
connect this knowledge tracing model with the PCG approach developed in our previous work~\cite{Valls2017grammar}. 

{\bf Acknowledgements.} This project is partially supported by Cyberlearning NSF grant 1523116.

\end{document}